\documentclass[10pt,twocolumn]{article}
\usepackage{simpleConference}
\usepackage{times}
\usepackage{graphicx}
\usepackage{amssymb}
\usepackage{url,hyperref}

\usepackage[normalem]{ulem}
\useunder{\uline}{\ul}{}

\begin{document}

\title{UAMD-Net: A Unified Adaptive Multimodal Neural Network for Dense Depth Completion}

\author{Guancheng Chen, Junli Lin, Huabiao Qin\thanks{Corresponding author.} \\
\\
South China University of Technology, Guangzhou, China \\
eechengc@mail.scut.edu.cn, 1070403885@qq.com, eehbqin@scut.edu.cn \\
}

\maketitle
\thispagestyle{empty}

\begin{abstract}
Depth prediction is a critical problem in robotics applications especially autonomous driving. Generally, depth prediction based on binocular stereo matching and fusion of monocular image and laser point cloud are two mainstream methods. However, the former usually suffers from overfitting while building cost volume, and the latter has a limited generalization due to the lack of geometric constraint. To solve these problems, we propose a novel multimodal neural network, namely \textbf{UAMD-Net}, for dense depth completion based on fusion of binocular stereo matching and the weak constrain from the sparse point clouds. Specifically, the sparse point clouds are converted to sparse depth map and sent to the \emph{multimodal feature encoder} (\textbf{MFE}) with binocular image, constructing a cross-modal cost volume. Then, it will be further processed by the \emph{multimodal feature aggregator} (\textbf{MFA}) and the depth regression layer. Furthermore, the existing multimodal methods ignore the problem of modal dependence, that is, the network will not work when a certain modal input has a problem. Therefore, we propose a new training strategy called \textbf{Modal-dropout} which enables the network to be adaptively trained with multiple modal inputs and inference with specific modal inputs. Benefiting from the flexible network structure and adaptive training method, our proposed network can realize unified training under various modal input conditions. Comprehensive experiments conducted on KITTI depth completion benchmark demonstrate that our method produces robust results and outperforms other state-of-the-art methods.
\end{abstract}

\section{Introduction}

Dense depth prediction is of great significance to the robotics applications such as autonomous driving. The acquisition of depth information is the prerequisite for solving the tasks like obstacle avoidance, 3D object detection and 3D scene reconstruction \cite{laga2020survey}. Typically, there are two major application environments, indoors and outdoors. For the former, the mainstream method is using the depth camera to proactively acquire the depth information or utilizing the stereo vision in a passive way. But for the latter, it is better to apply the stereo vision or LiDAR sensors \cite{tang2020learning}. Besides, estimating the depth directly from monocular image \cite{saxena2005learning, eigen2014depth, eigen2015predicting, wang2015towards, xu2017multi, fu2018deep} is also an attempt although it is a morbid problem. Recently, stereo vision algorithms have achieved an impressive progress both in supervised \cite{chang2018pyramid, guo2019group, zhang2020domain, cheng2020hierarchical} and in self-supervised way \cite{godard2017unsupervised, zhong2017self, pilzer2018unsupervised, luo2018single}, but the problems of weak texture failure and over fitting are still unsolved, which will lead to limited accuracy. In contrast, LiDAR sensors can provide reliable and accurate depth sensing. But unfortunately, current LiDAR sensors only acquire sparse depth measurements which is not sufficient for real applications such as robotic navigation. Therefore, how to get both dense and accurate depth perception is still a challenging topic.

\begin{figure*}[htp]
\centerline{\includegraphics[width=0.85\textwidth]{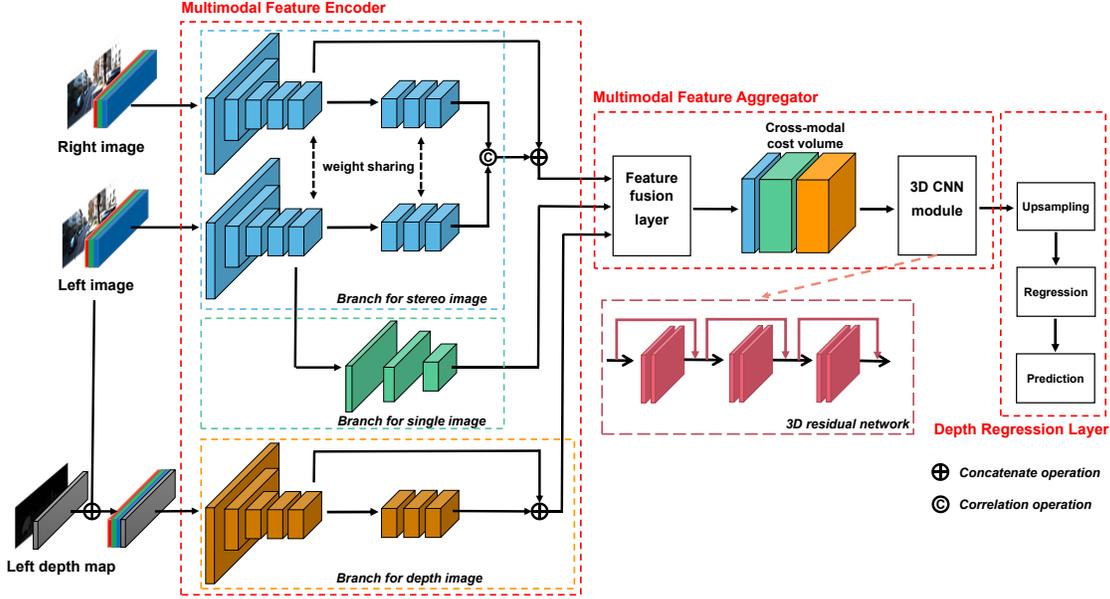}}
\caption{The network architecture of the proposed UAMD-Net, which consists of three main components: multimodal feature encoder (MFE), multimodal feature aggregator (MFA) and Depth Regression Layer (DRL).}
\label{fg1}
\end{figure*}

Many recent works on this topic turn on the trend of multimodal learning by fusing monocular image information and sparse depth measurements for depth completion. Cheng et al. \cite{cheng2019learning, cheng2020cspn++} proposed to utilize the convolutional spatial propagation network (CSPN, CSPN++) to assemble the features learning from monocular image and the corresponding LiDAR scans. Similarly, Park et al. \cite{park2020non} continued this idea and put forward the non-local spatial propagation network (NLSPN) that predicted non-local neighbors for each pixel, aiming to solve the mixed-depth problem. In contrast to the application of spatial propagation scheme, Tang et al. \cite{tang2020learning} proposed to fuse the LiDAR data and RGB image information by performing GuideNet which consists of learnable content-dependent and spatially-variant kernels. Zhao et al. \cite{zhao2021adaptive} proposed to adopt the graph propagation to capture the observed spatial contexts. More recently, PENet \cite{hu2021penet} and FCFR-Net \cite{liu2020fcfr} were proposed to carried out the depth completion through a two-stage coarse-to-fine mechanism. Although these methods have achieved remarkable results, its demand for a huge amount of label data and a long training time for convergence cannot be ignored. Besides, the hard fitting of monocular image and scene depth lacks geometric constraints, which will lead to scene dependence and limited generalization.

Instead of establishing the multimodal depth completion network based on monocular image and sparse point cloud, LiStereo \cite{zhang2020listereo} was the pioneer to make use of the multimodal learning of binocular image and sparse depth measurements. Except training on supervised mode, it also can be trained on semi-supervised mode benefited from the view synthesis scheme of binocular image and the weak constraint from the sparse point cloud. However, its feature fusion and aggregation modules are not sufficient for producing satisfactory results.

\begin{figure*}
\centerline{\includegraphics[width=0.85\textwidth]{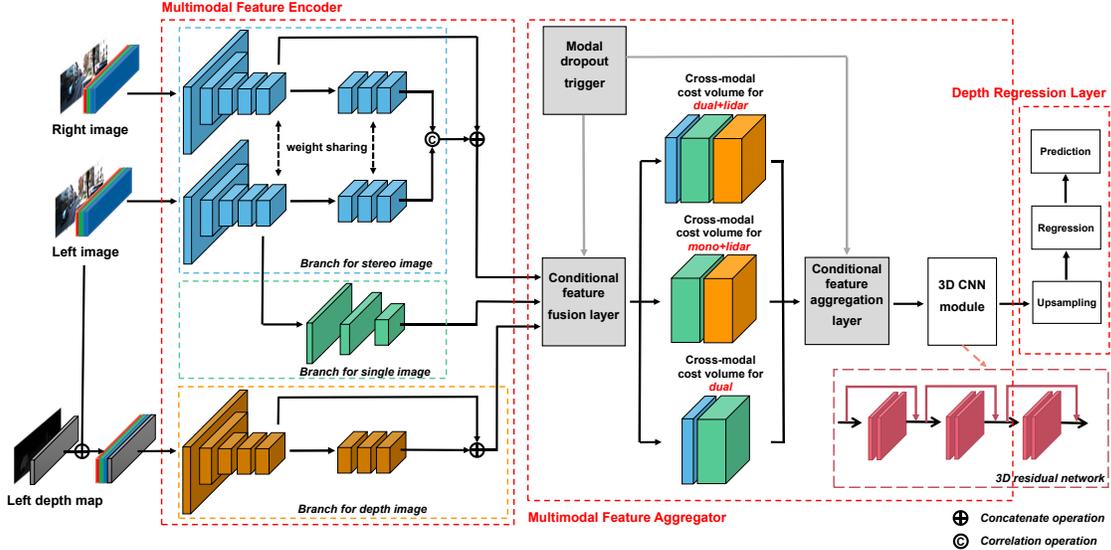}}
\caption{The extended network structure of UAMD-Net integrating three new modules including modal dropout trigger (MDT), conditional feature fusion layer (CFFL) and conditional feature aggregation layer (CFAL), which is designed for carrying out the proposed Modal-dropout training scheme.}
\label{fg2}
\end{figure*}

Besides, all these existing multimodal methods ignore the problem of modal dependence, which means the network will not work when a certain modal input has a problem.

To address the aforementioned issues, in this paper we propose a unified multimodal neural network, namely \textbf{UAMD-Net}, that is capable to fuse the feature learning of binocular image and sparse depth map. Specifically, it consists of the \textbf{MFE} and \textbf{MFA} module which can extract the cross-modal features to construct the 4D cost volume and then accomplish the feature aggregation based on 3D convolution. Besides, to solve the modal dependence problem, we propose a new training scheme called \textbf{Modal-dropout} which is capable to adaptively train the network with multiple modal inputs and inference with specific modal inputs. In particular, the flexible network structure and adaptive training method enable the network to realize unified training under various modal input conditions, including binocular stereo matching (\emph{\textbf{dual}}), fusion of monocular image and sparse depth map (\emph{\textbf{mono\_lidar}}), and combination of binocular image and sparse depth map (\emph{\textbf{dual\_lidar}}). We conducted extensive experiments on KITTI depth completion benchmark and the results show that our technique achieves strong results and outperforms current state-of-the art methods.

In short, the contributions of our research include:
\begin{itemize}
\item We propose a novel multimodal neural network for realizing depth completion, which we called \textbf{UAMD-Net}. It is capable to combine the advantages of binocular stereo matching and sparse point cloud constraint to get rid of the risk of over fitting and obtain better generalization performance.
\item We propose a new training strategy called \textbf{Modal-dropout} to solve the modal dependence problem for multimodal learning. To the best of our knowledge, this is the first trial to provide a viable solution to the modal dependence problem.
\item Our proposed network has great flexibility to realize unified training under various modal input conditions. Extensive experimental results on KITTI depth completion benchmark demonstrate the superiority of our proposed method quantificationally and qualitatively.
\end{itemize}

\section{Related Work}

\subsection{Monocular Depth Estimation}

The first work of monocular depth estimation can be traced back to 2005, when Saxena et al. constructed a Gaussian MRF probabilistic model by handcraft features to directly perform RGB-to-depth regression. With the popularity of convolutional neural network (CNN), a series of CNN-based monocular depth estimation networks have been proposed \cite{eigen2014depth, eigen2015predicting, wang2015towards, xu2017multi, fu2018deep, ma2019self}, and achieved a constant improvement in accuracy. However, since monocular depth estimation is an ill-conditioned problem that relies heavily on the learning of scene texture and structure information, it is difficult to solve the problem of scene generalization.

\subsection{Depth Estimation Based on Stereo Match}

Unlike monocular depth estimation, binocular depth estimation can use the geometric constraints of stereo matching for depth prediction. Generally, it can be divided into two learning paradigms. The first one is supervised learning. This kind of methods usually apply a two-stream neural network to extract binocular image features, then construct the matching cost volume for further depth regression, like \cite{chang2018pyramid, guo2019group, zhang2020domain, cheng2020hierarchical}. The second one is self-supervised learning. This kind of methods perform self-supervised learning through the mechanism of view synthesis without the need for labeled data, such as \cite{godard2017unsupervised, zhong2017self, pilzer2018unsupervised, luo2018single}. Although binocular depth estimation can utilize the geometric prior information of the scene, the constraint based on the similarity of the matching pixels in the left and right views is not strong, which easily leads to the problem of overfitting.

\subsection{ Depth Completion Based on Multimodal Learning}

Recently, depth completion based on multimodal learning receives raising spotlight. CSPN \cite{cheng2019learning} started a fashion of fusing of features learning from monocular image and the corresponding depth measurement through the convolutional spatial propagation network. Soon after, CSPN++ \cite{cheng2020cspn++} further improves the effectiveness and efficiency of CSPN by learning adaptive convolutional kernel sizes and the number of iterations for the propagation. Park et al. \cite{park2020non} put forward the non-local spatial propagation network (NLSPN) that predicted non-local neighbors for each pixel, aiming to solve the mixed-depth problem. Except the application of spatial propagation scheme, Tang et al. \cite{tang2020learning} proposed GuideNet to fuse the LiDAR data and RGB image information by performing learnable content-dependent and spatially-variant kernels.  Zhao et al. \cite{zhao2021adaptive} proposed to adopt the graph propagation to model the observed spatial contexts with depth values, so as to better guide the recovery of the unobserved pixels' depth. More recently, PENet \cite{hu2021penet} and FCFR-Net \cite{liu2020fcfr} were proposed to carried out the depth completion through a two-stage coarse-to-fine mechanism and achieved impressive results. However, these methods need be trained in supervised learning mode, which require a large amount of annotated label. In order to get rid of the limit of supervised learning, Zhang et al. proposed LiStereo \cite{zhang2020listereo} to realize depth completion by accomplishing the multimodal learning of binocular image and sparse depth measurements. It can be trained on semi-supervised mode based on the view synthesis scheme of binocular image and the weak constraint from the sparse point cloud. However, since the sparse point cloud is not able to offer enough constraint like the label data, and its feature fusion and aggregation modules are not sufficient enough, it still has a large performance gap compared with supervised learning methods. In this paper, we propose a novel multimodal neural network which tries to combine the advantages of binocular stereo matching and sparse point cloud constraint, aiming at improving the performance of both supervised and semi-supervised learning modes.

\begin{table*}[htbp]\small
\centering
\caption{Ablation study on weights of loss for semi-supervised learning. Our UAMD-Net is trained in semi-supervised mode with the modal input \emph{\textbf{dual\_lidar}}.}
\label{tb1}
\begin{tabular}{cccccccccc}
\hline
Loss weights & \begin{tabular}[c]{@{}c@{}}$w_l=1$\\ $w_p=0$\end{tabular} & \begin{tabular}[c]{@{}c@{}}$w_l=1$\\ $w_p=0.2$\end{tabular} & \begin{tabular}[c]{@{}c@{}}$w_l=1$\\ $w_p=0.4$\end{tabular} & \begin{tabular}[c]{@{}c@{}}$w_l=1$\\ $w_p=0.6$\end{tabular} & \begin{tabular}[c]{@{}c@{}}$w_l=1$\\ $w_p=0.8$\end{tabular} & \begin{tabular}[c]{@{}c@{}}$w_l=1$\\ $w_p=1.0$\end{tabular} & \begin{tabular}[c]{@{}c@{}}$w_l=1$\\ $w_p=1.3$\end{tabular} & \begin{tabular}[c]{@{}c@{}}$w_l=1$\\ $w_p=1.5$\end{tabular} & \begin{tabular}[c]{@{}c@{}}$w_l=0$\\ $w_p=1.0$\end{tabular} \\ \hline
RMSE (mm)     & 1725.587                                                  & 1513.612                                                    & 1373.387                                                    & 1337.512                                                    & 1270.628                                                    & 1305.006                                                    & \textbf{1267.047}                                           & 1381.939                                                    & 2587.277                                                    \\ \hline
\end{tabular}
\end{table*}

\begin{table}[htbp]\small
\centering
\caption{Ablation study on different learning mode for various modal input combinations.}
\label{tb2}
\begin{tabular}{ccccc}
\hline
                                  & \begin{tabular}[c]{@{}c@{}}iRMSE\\ (1/km)\end{tabular} & \begin{tabular}[c]{@{}c@{}}iMAE\\ (1/km)\end{tabular} & \begin{tabular}[c]{@{}c@{}}RMSE\\ (mm)\end{tabular} & \begin{tabular}[c]{@{}c@{}}MAE\\ (mm)\end{tabular} \\ \hline
\textit{\textbf{supervised}}      &                                                        &                                                       &                                                     &                                                    \\
dual\_lidar                        & 1.747                                                  & 1.007                                                 & \textbf{669.166}                                    & 252.580                                            \\
mono\_lidar                        & 1.938                                                  & 1.098                                                 & 918.067                                             & 346.224                                            \\
dual                              & 6.400                                                  & 4.545                                                 & 1163.147                                            & 603.036                                            \\ \hline
\textit{\textbf{semi-supervised}} &                                                        &                                                       &                                                     &                                                    \\
dual\_lidar                        & 4.954                                                  & 1.864                                                 & \textbf{1267.047}                                   & 460.880                                            \\
dual                              & 4.639                                                  & 2.031                                                 & 2587.277                                            & 731.396                                            \\ \hline
\end{tabular}
\end{table}

\begin{table*}[htbp]
\centering
\caption{Ablation study on Modal-dropout training scheme: both training and validating with various modal input combinations.}
\label{tb3}
\begin{tabular}{cccccc}
\hline
                                  & iRMSE (1/km)       & iMAE (1/km) & RMSE (mm)         & MAE (mm) & Performance attenuation \\ \hline
\textit{\textbf{supervised}}      &                    &             &                   &          &                         \\
dual\_lidar                       & 1.873              & 1.095       & \textbf{730.327}  & 284.489  & $\downarrow$9.14\%      \\
mono\_lidar                       & 9.423              & 5.986       & 1306.826          & 610.729  & $\downarrow$12.35\%     \\
dual                              & 2.086              & 1.234       & 973.540           & 385.330  & $\downarrow$6.04\%      \\
\textit{avg.}                     & -                  & -           & -                 & -        & $\downarrow$9.18\%      \\ \hline
\textit{\textbf{semi-supervised}} & \textit{\textbf{}} &             &                   &          &                         \\
dual\_lidar                       & 4.269              & 1.862       & \textbf{1419.445} & 515.308  & $\downarrow$13.02\%     \\
dual                              & 4.343              & 2.014       & 3087.818          & 744.682  & $\downarrow$19.35\%     \\
\textit{avg.}                     & -                  & -           & -                 & -        & $\downarrow$16.19\%     \\ \hline
\end{tabular}
\end{table*}

\begin{table*}[htbp]
\centering
\caption{Ablation study on Modal-dropout training scheme: training with various modal input combinations while validating with specific modal input combination. For example, the first section means training with \{dual\_lidar, mono\_lidar, dual\} while validating with dual\_lidar.}
\label{tb4}
\begin{tabular}{cccccc}
\hline
                                  & iRMSE (1/km)         & iMAE (1/km)          & RMSE (mm)            & MAE (mm)             & Performance attenuation \\ \hline
\textit{\textbf{supervised}}      &                      &                      &                      &                      &                         \\
\textit{\textbf{dual\_lidar}}     &                      &                      &                      &                      &                         \\
dual\_lidar                       & 1.888                & 1.065                & \textbf{747.816}     & 285.691              & $\downarrow$11.75\%     \\
mono\_lidar                       & 7.135                & 4.908                & 1319.096             & 631.323              & $\downarrow$13.48\%     \\
dual                              & 1.165                & 2.035                & 991.919              & 373.009              & $\downarrow$8.04\%      \\
\textit{avg.}                     & -                    & -                    & -                    & -                    & $\downarrow$11.09\%     \\
\textit{\textbf{mono\_lidar}}     &                      &                      &                      &                      &                         \\
dual\_lidar                       & 1.810                & 1.036                & \textbf{709.975}     & 269.336              & $\downarrow$6.10\%      \\
mono\_lidar                       & 4.758                & 3.011                & 1319.360             & 526.489              & $\downarrow$13.43\%     \\
dual                              & 2.054                & 1.139                & 967.731              & 364.451              & $\downarrow$5.41\%      \\
\textit{avg.}                     & -                    & -                    & -                    & -                    & \textbf{$\downarrow$8.31\%}      \\
\textit{\textbf{dual}}            &                      &                      &                      &                      &                         \\
dual\_lidar                       & 1.888                & 1.056                & \textbf{731.881}     & 278.886              & $\downarrow$9.37\%      \\
mono\_lidar                       & 6.138                & 3.760                & 1369.923             & 527.990              & $\downarrow$17.78\%     \\
dual                              & 2.130                & 1.154                & 988.307              & 370.867              & $\downarrow$7.65\%      \\
\textit{avg.}                     & -                    & -                    & -                    & -                    & $\downarrow$11.60\%     \\ \hline
\textit{\textbf{semi-supervised}} & \textit{\textbf{}}   &                      &                      &                      &                         \\
\textit{\textbf{dual\_lidar}}     & \multicolumn{1}{l}{} & \multicolumn{1}{l}{} & \multicolumn{1}{l}{} & \multicolumn{1}{l}{} & \multicolumn{1}{l}{}    \\
dual\_lidar                       & 5.129                & 1.887                & \textbf{1320.503}    & 492.800              & $\downarrow$4.22\%      \\
dual                              & 5.692                & 2.156                & 7733.671             & 955.438              & -                       \\
\textit{avg.}                     & -                    & -                    & -                    & -                    & -                       \\
\textit{\textbf{dual}}            & \multicolumn{1}{l}{} & \multicolumn{1}{l}{} & \multicolumn{1}{l}{} & \multicolumn{1}{l}{} & \multicolumn{1}{l}{}    \\
dual\_lidar                       & 4.663                & 1.811                & \textbf{1351.973}    & 488.343              & $\downarrow$6.70\%      \\
dual                              & 4.140                & 2.020                & 2902.286             & 742.696              & $\downarrow$12.18\%     \\
\textit{avg.}                     & -                    & -                    & -                    & -                    & $\downarrow$9.44\%      \\ \hline
\end{tabular}
\end{table*}

\begin{table}[htbp]\scriptsize
\centering
\caption{Ablation study on noise label learning. Our UAMD-Net is trained in semi-supervised mode with the modal input \emph{\textbf{dual\_lidar}}.}
\label{tb5}
\begin{tabular}{cccccc}
\hline
$w_n$         & 0        & 0.1               & 0.2      & 0.3      & 0.4       \\ \hline
RMSE (mm)     & 1267.05  & \textbf{1225.85}  & 1270.06  & 1264.00  & 1269.72   \\ \hline
\end{tabular}
\end{table}

\begin{table*}[htbp]\scriptsize
\centering
\caption{Robustness against different modal failure situations. In order to deal with the situations of image failure, our UAMD-Net switches to work with the modal input \emph{\textbf{mono\_lidar}}.}
\label{tb6}
\begin{tabular}{c|cccc|cccc|cccc}
\hline
\multicolumn{1}{l|}{}                                          & \multicolumn{4}{c|}{image failure (half\_h)}                                                                                                                                                                              & \multicolumn{4}{c|}{image failure (half\_v)}                                                                                                                                                                              & \multicolumn{4}{c}{image failure (full)}                                                                                                                                                                                   \\ \cline{2-13}
                                                               & \begin{tabular}[c]{@{}c@{}}iRMSE\\ (1/km)\end{tabular} & \begin{tabular}[c]{@{}c@{}}iMAE\\ (1/km)\end{tabular} & \begin{tabular}[c]{@{}c@{}}RMSE\\ (mm)\end{tabular} & \begin{tabular}[c]{@{}c@{}}MAE\\ (mm)\end{tabular} & \begin{tabular}[c]{@{}c@{}}iRMSE\\ (1/km)\end{tabular} & \begin{tabular}[c]{@{}c@{}}iMAE\\ (1/km)\end{tabular} & \begin{tabular}[c]{@{}c@{}}RMSE\\ (mm)\end{tabular} & \begin{tabular}[c]{@{}c@{}}MAE\\ (mm)\end{tabular} & \begin{tabular}[c]{@{}c@{}}iRMSE\\ (1/km)\end{tabular} & \begin{tabular}[c]{@{}c@{}}iMAE\\ (1/km)\end{tabular} & \begin{tabular}[c]{@{}c@{}}RMSE\\ (mm)\end{tabular} & \begin{tabular}[c]{@{}c@{}}MAE\\ (mm)\end{tabular} \\ \hline
PENet \cite{hu2021penet}                                       & 4.448                                                  & 1.626                                                 & 1630.624                                            & 463.079                                            & 3.545                                                  & 1.320                                                 & 1779.584                                            & 421.798                                            & 5.173                                                  & 1.854                                                 & 2531.038                                            & 656.476                                            \\
ACMNet \cite{zhao2021adaptive}                        & 3.255                                                  & 1.086                                                 & \textbf{1131.010}                                   & 274.131                                            & 2.592                                                  & 0.971                                                 & \textbf{1151.714}                                   & 261.263                                            & 3.307                                                  & 1.111                                                 & {\ul 1439.257}                                      & 319.110                                            \\
\begin{tabular}[c]{@{}c@{}}UAMD-Net\\ (mono\_lidar)\end{tabular} & 4.821                                                  & 3.009                                                 & {\ul 1348.657}                                      & 525.945                                            & 4.821                                                  & 3.009                                                 & {\ul 1348.657}                                      & 525.945                                            & 4.821                                                  & 3.009                                                 & \textbf{1348.657}                                   & 525.945                                            \\ \hline
\end{tabular}
\end{table*}

\begin{table*}[htbp]
\centering
\caption{Robustness against different modal failure situations. In order to deal with the situations of rotation failure and LiDAR data failure, our UAMD-Net switches to work with the modal input \emph{\textbf{dual}}.}
\label{tb7}
\begin{tabular}{c|cccc|cccc}
\hline
\multicolumn{1}{l|}{} & \multicolumn{4}{c|}{rotation failure}                                                                                                                                                                                     & \multicolumn{4}{c}{LiDAR data failure}                                                                                                                                                                                    \\ \cline{2-9}
                      & \begin{tabular}[c]{@{}c@{}}iRMSE\\ (1/km)\end{tabular} & \begin{tabular}[c]{@{}c@{}}iMAE\\ (1/km)\end{tabular} & \begin{tabular}[c]{@{}c@{}}RMSE\\ (mm)\end{tabular} & \begin{tabular}[c]{@{}c@{}}MAE\\ (mm)\end{tabular} & \begin{tabular}[c]{@{}c@{}}iRMSE\\ (1/km)\end{tabular} & \begin{tabular}[c]{@{}c@{}}iMAE\\ (1/km)\end{tabular} & \begin{tabular}[c]{@{}c@{}}RMSE\\ (mm)\end{tabular} & \begin{tabular}[c]{@{}c@{}}MAE\\ (mm)\end{tabular} \\ \hline
PENet \cite{hu2021penet}                & 524.298                                                & 430.731                                               & 17517.829                                           & 12812.575                                          & -                                                      & -                                                     & -                                                   & -                                                  \\
ACMNet \cite{zhao2021adaptive}               & 276.951                                                & 247.391                                               & {\ul 16856.885}                                     & 12277.822                                          & -                                                      & -                                                     & -                                                   & -                                                  \\
UAMD-Net (dual)         & 2.115                                                  & 1.151                                                 & \textbf{956.168}                                    & 361.121                                            & 2.115                                                  & 1.151                                                 & \textbf{956.168}                                    & 361.121                                            \\ \hline
\end{tabular}
\end{table*}

\begin{table}[htbp]\small
\centering
\caption{Comparison with other supervised methods. Our UAMD-Net is trained in supervised mode with the modal input \emph{\textbf{dual\_lidar}}.}
\label{tb8}
\begin{tabular}{ccccc}
\hline
              & \begin{tabular}[c]{@{}c@{}}iRMSE\\ (1/km)\end{tabular} & \begin{tabular}[c]{@{}c@{}}iMAE\\ (1/km)\end{tabular} & \begin{tabular}[c]{@{}c@{}}RMSE\\ (mm)\end{tabular} & \begin{tabular}[c]{@{}c@{}}MAE\\ (mm)\end{tabular} \\ \hline
PENet \cite{hu2021penet}         & 2.159                                                  & 0.903                                                 & 756.667                                             & 209.369                                            \\
ACMNet \cite{zhao2021adaptive}        & 2.099                                                  & 0.868                                                 & 765.210                                             & 205.503                                            \\
GuideNet \cite{tang2020learning}      & -                                                      & -                                                     & 763.3                                               & -                                                  \\
NLSPN \cite{park2020non}         & 2.0                                                    & 0.8                                                   & 776.3                                               & 198.5                                              \\
UAMD-Net (ours) & 1.729                                                  & 0.999                                                 & \textbf{677.132}                                    & 254.056                                            \\ \hline
\end{tabular}
\end{table}

\begin{table}[htbp]\small
\centering
\caption{Comparison with other semi-supervised methods. Our UAMD-Net is trained in semi-supervised mode with the modal input \emph{\textbf{dual\_lidar}}.}
\label{tb9}
\begin{tabular}{ccccc}
\hline
                & \begin{tabular}[c]{@{}c@{}}iRMSE\\ (1/km)\end{tabular} & \begin{tabular}[c]{@{}c@{}}iMAE\\ (1/km)\end{tabular} & \begin{tabular}[c]{@{}c@{}}RMSE\\ (mm)\end{tabular} & \begin{tabular}[c]{@{}c@{}}MAE\\ (mm)\end{tabular} \\ \hline
Sparse2dense \cite{ma2019self}    & 4.08                                                   & 1.61                                                  & 1301.05                                             & 352.22                                             \\
LiStereo \cite{zhang2020listereo}        & 3.84                                                   & 1.32                                                  & 1278.87                                             & 326.10                                             \\
UAMD-Net (ours) & 4.71                                                   & 1.82                                                  & \textbf{1241.10}                                    & 464.38                                             \\ \hline
\end{tabular}
\end{table}

\section{Proposed Method}

In this section, we firstly describe the network architecture of the proposed \textbf{UAMD-Net}, shown in Fig. \ref{fg1}, which is mainly divided into three components: \textbf{MFE}, \textbf{MFA} and \textbf{DRL}. Next, we detail the proposed adaptive multimodal training strategy \textbf{Modal-dropout} and the extended network structure of \textbf{UAMD-Net} which is designed for unified training under various modal input conditions. Finally, we introduce the objective function design.

\subsection{Multimodal Neural Network for Depth Completion}

\textbf{MFE:} For multimodal inputs, we design three branches to extract the specific modal features. The branch for stereo image and the branch for depth map have the same configuration, consisting of multiple convolutional layers with ReLUs as the activation functions. We use the branch for stereo image to extract the features from both left and right image, and then obtain the cross-modal features by accomplishing the correlation operation. Besides, we concatenate the image features from the middle layer to enhance the feature representation ability. We use the branch for depth map to extract the cross-modal features from the concatenation of monocular image and the corresponding sparse depth map. Moreover, we design a branch for single image in order to enhance the features from the image domain. More specific settings are specified in the supplementary material.

\textbf{MFA:} After acquiring the multimodal features from different branches, we design a feature fusion layer to construct the cross-modal 4D cost volume by fusing the multimodal features. Then, inspired by \cite{chang2018pyramid}, we establish a simple yet effective 3D CNN module by stacking six $3\times3\times3$ 3D convolutional layers and three residual blocks, which can aggregate the features from both spatial and channel dimensions.

\textbf{DRL:} After feature aggregation, we apply the trilinear interpolation on the disparity feature map to recover the resolution to $H\times W\times D$ ($H, W$ represent the height and width of the image, $D$ denotes the disparity range which we set as 192). Then, we adopt the softmax operation to carry out the disparity regression. In this case, the features in $D$ dimension are considered to be the probability of the corresponding disparity. Finally, the disparity map will be transformed to depth map according to the stereo constraint: $d=fl/disp$, where $d$ denotes the depth map, $b$ denotes the length of baseline, $fl$ denotes the focal length of the camera, and $disp$ denotes the predicted disparity map.

\subsection{Adaptive Multimodal Training Strategy and the Extended Unified Network Structure}

We get inspiration from the universal training method \textbf{Dropout}, which randomly discards nodes to prevent network overfitting. Similarly, we propose to randomly drop the specific modal inputs during training while inference with fixed modal inputs, so as to prevent the network from being limited to specific modal inputs, which solves the modal dependence problem. Naturally, we name this training strategy \textbf{Modal-dropout}. To carry out this training scheme, we need to further extend the network structure. As shown in Fig. \ref{fg2}, we design the \textbf{MDT} component to guide the \textbf{CFFL} and the \textbf{CFAL} to adaptively accomplish the feature fusion and aggregation, respectively.

\textbf{MDT:} The key of \textbf{MDT} is to realize the random sampling of three modal input combinations: \emph{\textbf{dual\_lidar}}, \emph{\textbf{mono\_lidar}}, and \emph{\textbf{dual}}, which can be formulated as follow:
\begin{equation}\label{eq1}
X \sim P\left\{ {X = k} \right\} = 1/3,k = 1,2,3
\end{equation}
where $X$ denotes the sample variable of three cases.

Therefore, the formulation of the network trained on supervised mode can be described as:
\begin{equation}\label{eq2}
\left\{ {\begin{array}{*{20}{c}}
{\begin{array}{*{20}{c}}
{d = f\left( {{I_l},{I_r},{D_l}} \right)}&{X = 1}
\end{array}}\\
{\begin{array}{*{20}{c}}
{d = f\left( {{I_l},{D_l}} \right)}&{X = 2}
\end{array}}\\
{\begin{array}{*{20}{c}}
{d = f\left( {{I_l},{I_r}} \right)}&{X = 3}
\end{array}}
\end{array}} \right.
\end{equation}
and the formulation of the network trained on semi-supervised mode can be described as:
\begin{equation}\label{eq3}
\left\{ {\begin{array}{*{20}{c}}
{\begin{array}{*{20}{c}}
{d = f\left( {{I_l},{I_r},{D_l},{D_r}} \right)}&{X = 1}
\end{array}}\\
{\begin{array}{*{20}{c}}
{d = f\left( {{I_l},{I_r}} \right)}&{X = 2}
\end{array}}
\end{array}} \right.
\end{equation}
where $d$ denotes the predicted depth map, $f$ denotes the network, $I_l$, $I_r$, $D_l$, $D_r$ denote the left image, right image, left sparse depth map, right sparse depth map, respectively.

\textbf{CFFL:} To realize the adaptive multimodal training, we construct a \emph{conditional feature fusion layer}. It receives the command from the \textbf{MDT} to adaptively produce three forms of 4D cost volume, corresponding to three different modal input combinations: \emph{\textbf{dual\_lidar}}, \emph{\textbf{mono\_lidar}}, and \emph{\textbf{dual}}.

\textbf{CFAL:} Since the feature dimension of 4D cost volume is changeable, we establish a \emph{conditional feature aggregation layer} to adaptively accomplish the feature aggregation. It receives the same command as the \textbf{CFFL} from the \textbf{MDT}. We realize it by constructing two 3D convolutional layers with ReLUs as the activation functions. The input feature dimension is variant while the output feature dimension is fixed.

\subsection{Objective Function Design}

Our network can be trained in both supervised and semi-supervised mode benefited from the view synthesis scheme of stereo vision. For supervised learning, we optimize \textbf{UAMD-Net} by minimizing the following $L_2$ loss,
\begin{equation}\label{eq4}
Los{s_{\sup }} = \frac{1}{N}\sum\limits_{p \in {P_v}} {{{\left\| {d_p^{gt} - {d_p}} \right\|}^2}}
\end{equation}
where $P_v$ represents the set of the valid pixels. $d^{gt}_p$ and $d_p$ denote the ground truth and predicted depth at the pixel $p$, respectively. $N$ is the number of valid pixels.

For semi-supervised learning, we only provide sparse ground truth depth map for supervision. Since the density of sparse depth map is low and $L_2$ is more sensitive to the outliers, we adopt the following $L_1$ loss.
\begin{equation}\label{eq5}
Los{s_{lidar}} = \frac{1}{N}\sum\limits_{p \in {P_v}} {\left\| {d_p^{sp} - {d_p}} \right\|}
\end{equation}
where $d^{sp}_p$ denotes the ground truth sparse depth map.

Besides, we follow \cite{chang2018pyramid} to use a combination of an $L_1$ and single scale SSIM term as our photometric image reconstruction loss, which compares the input image $I^r_p$ and its reconstruction $\tilde{I^r_p}$.
\begin{footnotesize}
\begin{equation}\label{eq6}
Los{s_{photometric}} = \frac{1}{N}\sum\limits_{p \in {P_v}} {\alpha  \cdot SSIM\left( {I_p^r,\mathop {I_p^r}\limits^ \sim  } \right)}  + \left( {1 - \alpha } \right) \cdot \left\| {I_p^r - \mathop {I_p^r}\limits^ \sim  } \right\|
\end{equation}
\end{footnotesize}
Here, we use a simplified SSIM with a $3\times3$ block filter and set $\alpha=0.85$ . We noted that the correlation of stereo feature map can construct the cost volume for both left and right disparity, so in semi-supervised training, we construct the cost volume for right disparity with the reconstruction constraint of right image, while construct the cost volume for left disparity in the inference. Moreover, we apply the following objective for encouraging the depth to be locally smooth with an $L_1$ penalty on the depth gradients $\partial{}_d$.
\begin{small}
\begin{equation}\label{eq7}
Los{s_{gradient}} = \frac{1}{N}\sum\limits_{p \in {P_v}} {\left| {{\partial _x}d_p^r} \right|}  \cdot {e^{ - \left\| {{\partial _x}I_p^r} \right\|}} + \left| {{\partial _y}d_p^r} \right| \cdot {e^{ - \left\| {{\partial _y}I_p^r} \right\|}}
\end{equation}
\end{small}

Furthermore, inspired by \cite{zhu2020edge, tonioni2019unsupervised}, we employ the noise label learning strategy. We generate the noise depth map by the traditional depth estimation method Semi-Global Matching (SGM) for convenience. Since the accuracy of the noise depth map is limited and $L_2$ is more sensitive to the outliers, we adopt the following $L_1$ loss.
\begin{equation}\label{eq8}
Los{s_{noise}} = \frac{1}{N}\sum\limits_{p \in {P_v}} {\left\| {d_p^n - {d_p}} \right\|}
\end{equation}
where $d^n_p$ denotes the generated noise depth map.

Finally, the above objective will be combined to train in a multi-task learning fashion. $w_l$, $w_p$, $w_g$, $w_n$ represent the corresponding weight parameters, which will be fine-tuned according to the training feedback.
\begin{equation}\label{eq9}
\begin{array}{l}
Los{s_{semi}} = {w_l} \cdot Los{s_{lidar}} + {w_p} \cdot Los{s_{photometric}}\\
{\rm{                                  }} + {w_g} \cdot Los{s_{gradient}} + {w_n} \cdot Los{s_{noise}}
\end{array}
\end{equation}

\begin{figure*}
\centerline{\includegraphics[width=0.85\textwidth]{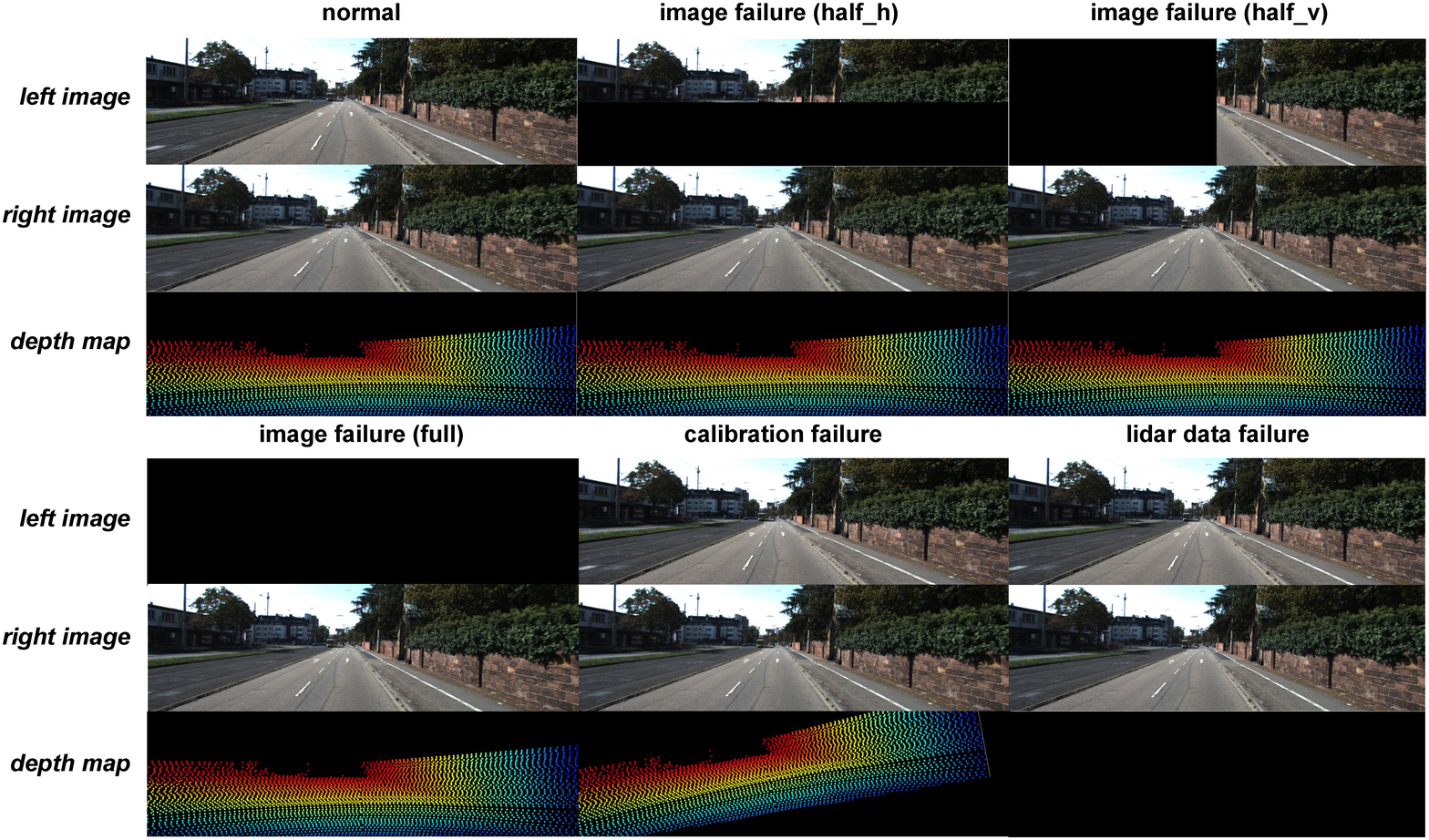}}
\caption{Modal failure situations: image failure (three forms including half of horizontal pixels, half of the vertical pixels and full pixels), camera and LiDAR calibration failure, and LiDAR data failure.}
\label{fg3}
\end{figure*}

\begin{figure*}
\centerline{\includegraphics[width=0.85\textwidth]{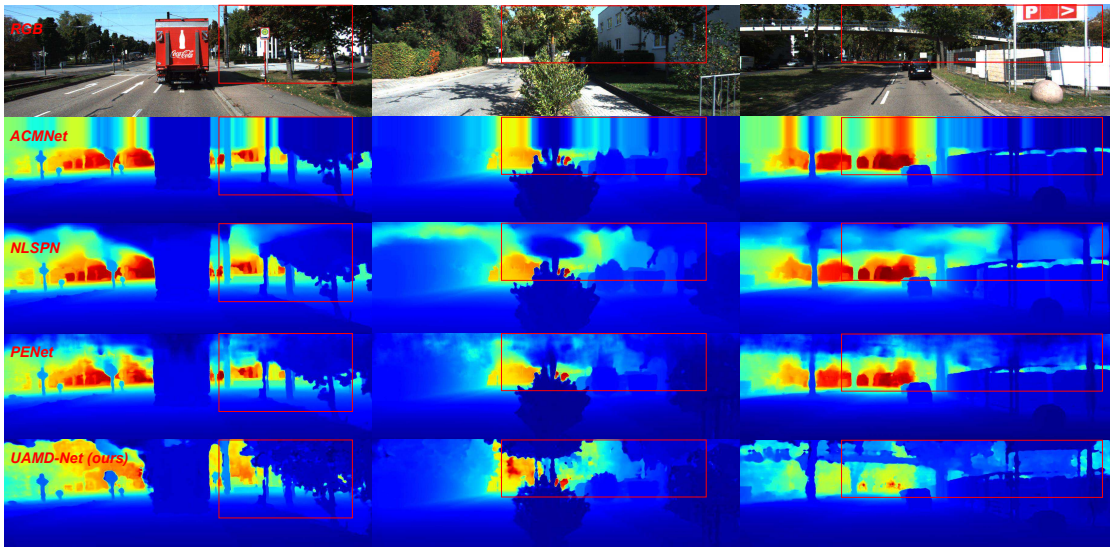}}
\caption{Qualitative results of different methods. From top to down are the input images, results of ACMNet \cite{zhao2021adaptive}, NLSPN \cite{park2020non}, PENet \cite{hu2021penet} and our \textbf{UAMD-Net} respectively.}
\label{fg4}
\end{figure*}

\section{Experiments}

\subsection{Experiments Settings}

\textbf{Benchmark Dataset:} The KITTI depth completion dataset \cite{geiger2012we} contains 42949 pair depth maps for training, 3426 pair depth maps for validation, and 1000 frames for testing. Specially a selection set of 1000 frames is also provided. These ground truth depth maps are generated by registering 11 LiDAR scans temporally and further refined with the corresponding stereo image pairs. Since the test set does not contain stereo images, we split the validation set into two sub-sets, 2426 pairs of stereo images for testing (split-test) and 1000 for validation (split-val) according to the selection set, which guarantees the fairness of the experimental comparison with other methods.

\textbf{Evaluation Metrics:} We follow the KITTI benchmark and exiting methods \cite{tang2020learning, zhao2021adaptive, hu2021penet} to use four standard metrics for evaluation: root mean squared error (RMSE), mean absolute error (MAE), root mean squared error of the inverse depth (iRMSE) and mean absolute error of the inverse depth (iMAE). Among them, RMSE and MAE measure the depth accuracy directly, while iRMSE and iMAE compute the mean error of the inverse depth, giving less weight to the far-away points. Since RMSE is more sensitive to the outliers, it is chosen as the dominant metric to rank the submissions on the KITTI leaderboard.

\textbf{Implementation Details:} Our network is implemented using PyTorch framework. For supervised learning, the learning rate begins at 1e-4 and is decayed by 0.5 at 10 epochs, 0.1 at 14 epochs and 0.01 at 17 epochs. For semi-supervised learning, the learning rate begins at 1e-4 and is decayed by 0.1 at 10e3 iterations, 0.01 at 14e3 iterations and 0.01 at 16e3 iterations. The batch size is set to 4 for training on 2 NVIDIA GTX 1080Ti GPUs for all models. We report the experimental results based on the validation set (split-val) for ablation studies of our proposed method. While compared with other start-of-the-art methods, we report the experimental results conducted on the test set (split-test). In all tables, bold values indicate the best performance, underlined values indicate the suboptimal performance.

\subsection{Ablation Study on Weight of Loss for Semi-supervised Learning}

The photometric loss is essential for the model to be trained in a self-supervised manner, so we investigate the influence of the weight of the photometric loss $w_p$. According to Eq. \ref{eq9}, we keep $w_l = 1$, $w_g = 0.01$, $w_n = 0$, and set $w_p$ between 0 to 1.5, and the results are shown in Table \ref{tb1}. It is clear that the photometric loss does help complete sparse depth map, reducing RMSE from 1725mm to 1267mm with $w_p = 0$ and $w_p = 1.3$ respectively. However, too much weight on photometric loss worsens the results. So we set $w_p = 1.3$ throughout the experiments.

\subsection{Ablation Study on Different Learning Modes for Various Modal Input Combinations}

In this section, we report the performance of our network trained with various modal input combinations in supervised and semi-supervised mode. As shown in Table \ref{tb2}, \emph{\textbf{dual\_lidar}} achieves the lowest RMSE while \emph{\textbf{dual}} acquires the highest in both supervised and semi-supervised mode, proving that our network can solve the problem of overfitting well. Besides, \emph{\textbf{mono\_lidar}} has a better performance than \emph{\textbf{dual}} benefited from the constraint of sparse point cloud.

\subsection{Ablation Study on Two Different Modal-out Training Schemes}

In this section, we study the performance of our two proposed \textbf{Modal-dropout} training strategies. The first one is both training and validating with various modal input combinations, the results are shown in Table \ref{tb3}. The second one is training with various modal input combinations while validating with specific modal input combination, the results are shown in Table \ref{tb4}. It is obvious that the model performance drops while trained with the \textbf{Modal-dropout} scheme, with the lowest average $8.31\%$. However, the model then will obtain the ability to inference with different modal input combinations, solving the modal dependence problem, which greatly improves its robustness against modal input failure situations.

\subsection{Ablation Study on Noise Label Learning}

We introduce the noise label learning scheme to further improve the performance of semi-supervised learning. As shown in Table \ref{tb5}, an appropriate proportion of noise labels is conducive to improve the performance, reducing RMSE from 1267 to 1225 with $w_n = 0.1$.

\subsection{Robustness against Different Modal Failure Situations}

In order to prove the effectiveness of our proposed \textbf{Modal-dropout} training strategy, we simulate the situations when the input modalities are problematic: image failure (three forms including half of horizontal pixels, half of the vertical pixels and full pixels), camera and LiDAR calibration failure, and LiDAR data failure, as shown in Fig. \ref{fg3}. The experimental results reported in Table \ref{tb6} and \ref{tb7} shows that current multimodal depth completion models like PENet \cite{hu2021penet} and ACMNet \cite{zhao2021adaptive} will have a large performance penalty for the case of image failure. Besides, they will complete failure for the case of calibration failure and LiDAR data failure. On the contrary, our model can switch to proper inference mode to maintain stable performance, demonstrating the great robustness of our method against modal input failure situations. More discussions are presented in the supplementary material.

\subsection{Comparison with State-of-the-Arts}

As shown in Table \ref{tb8} and \ref{tb9}, our method surpasses other state-of-the-art methods no matter in supervised or semi-supervised learning fashion. Especially, our method achieves an 80mm RMSE gap with the closest competitor, PENet \cite{hu2021penet}.

\subsection{Qualitative Results}

The qualitative results are reported in Fig. \ref{fg4}. From top to down are the input images, results of ACMNet \cite{zhao2021adaptive}, NLSPN \cite{park2020non}, PENet \cite{hu2021penet} and our \textbf{UAMD-Net} respectively. It shows that the depth map predicted by our method has great advantages in preserving edge details.

\section{Conclusion}

In this paper, we propose a unified multimodal neural network called \textbf{UAMD-Net} for depth completion task, aiming to combine the advantages of binocular stereo matching and sparse point cloud constraint to get rid of the risk of over fitting and obtain better generalization performance. Besides, to address the modal dependence problem, we further propose a new training strategy named \textbf{Modal-dropout}. The flexible network structure and adaptive training strategy enable the network realize unified training under various modal input conditions, which greatly improves the robustness of the multimodal neural network in the case of a modal input failure. Our experimental results demonstrate that the proposed method not only overcomes the modal dependence problem but also achieve better quantitative and qualitative performance compared with state-of-the-art methods.

\bibliographystyle{abbrv}
\bibliography{ref_for_acmmm}
\end{document}